\documentclass[11pt, a4paper, logo, copyright]{googledeepmind}

\usepackage[authoryear, sort&compress, round]{natbib}
\usepackage[utf8]{inputenc} 
\usepackage[T1]{fontenc}    
\usepackage{hyperref}       
\usepackage{url}            
\usepackage{booktabs}       
\usepackage{amsfonts}       
\usepackage{nicefrac}       
\usepackage{microtype}      
\usepackage{xcolor}         
\usepackage{arydshln}
\usepackage{graphicx}
\usepackage{subfigure}
\usepackage{algorithm}
\usepackage{algorithmic}
\usepackage{wrapfig}
\usepackage{amsmath}
\usepackage{caption}

 \usepackage[capitalise]{cleveref}
\usepackage[dvipsnames]{xcolor}

\title{Continual Learning in Vision-Language Models via Aligned Model Merging}

\author[1]{Ghada Sokar}
\author[1]{Gintare Karolina Dziugaite}
\author[1]{Anurag Arnab}
\author[1]{Ahmet Iscen}
\author[1]{Pablo Samuel Castro}
\author[1]{Cordelia Schmid}
\affil[1]{Google DeepMind}

\begin{abstract}
Continual learning is conventionally tackled through sequential fine-tuning, a process that,  while enabling adaptation, inherently favors plasticity over the stability needed to retain prior knowledge. While existing approaches attempt to mitigate catastrophic forgetting, a bias towards recent tasks persists as they build upon this sequential nature. In this work we present a new perspective based on model merging to maintain stability while still retaining plasticity. Rather than just sequentially updating the model weights, we propose merging newly trained task parameters with previously learned ones, promoting a better balance. To maximize the effectiveness of the merging process, we propose a simple mechanism that promotes learning aligned weights with previous ones, thereby avoiding interference when merging. We evaluate this approach on large Vision-Language Models (VLMs), and demonstrate its effectiveness in reducing forgetting, increasing robustness to various task orders and similarities, and improving generalization. 
\end{abstract}

\begin{document}

\maketitle

\section{Introduction}
\label{sec:intro}
Vision-language models (VLMs) have achieved remarkable success by leveraging pretraining on large-scale datasets \citep{devlin2018bert,radford2021learning}. 
Fine-tuning these pretrained models on specific downstream tasks yields task-specific variants with improved performance.
This approach, however, while effective for single-task optimization, falters in continual learning (CL) scenarios with a stream of diverse downstream tasks. Indeed, continual learning with large models poses two main challenges: (1) the model size, often reaching billions of parameters, can be prohibitively expensive for continual full fine-tuning; and (2) fine-tuning on a new task often leads to catastrophic forgetting \citep{robins1995catastrophic}, evidenced through performance degradation on previous tasks.

Parameter-efficient fine-tuning (PEFT) methods have emerged as an effective solution for addressing the first challenge
-- the prohibitive cost of continually fine-tuning large models -- primarily in single-task settings \citep{houlsby2019parameter,hulora,xu2023parameter}. By updating only a small subset of pre-trained parameters or introducing a limited number of new trainable parameters, PEFT achieves accuracy comparable to full fine-tuning. This approach not only enables more computationally efficient training but also helps mitigate overfitting, which is often associated with fine-tuning large models on small datasets. 
However, as PEFT was originally developed for single-task fine-tuning, its direct application in continual learning scenarios often leads to challenges, particularly catastrophic forgetting. Nevertheless, recent research has begun exploring the potential of specific PEFT techniques, notably Low-Rank Adaptation (LoRA) \citep{wang2023orthogonal,liu2024learning,yu2024boosting}, for continual learning.
LoRA introduces trainable low-rank matrices to a model, and optimizes these matrices on downstream tasks. Despite being computationally efficient, previous approaches still face difficulties such as scalability due to continuous accumulation of parameters, the need for task identifiers at test time, and catastrophic forgetting.

\begin{figure}
\vskip 0.2in
\begin{center}
\includegraphics[width=0.45\columnwidth]{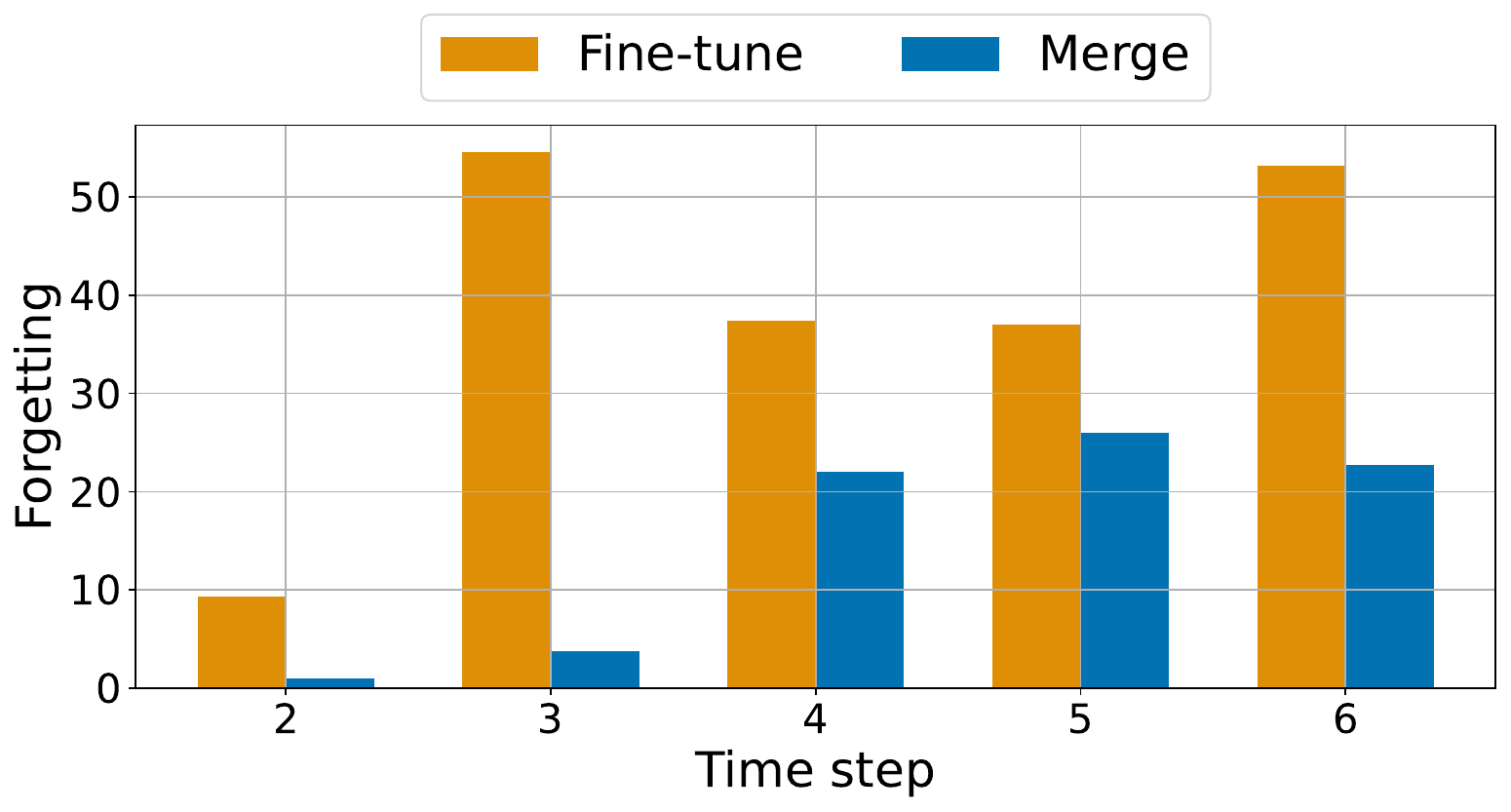}
\includegraphics[width=0.45\columnwidth]{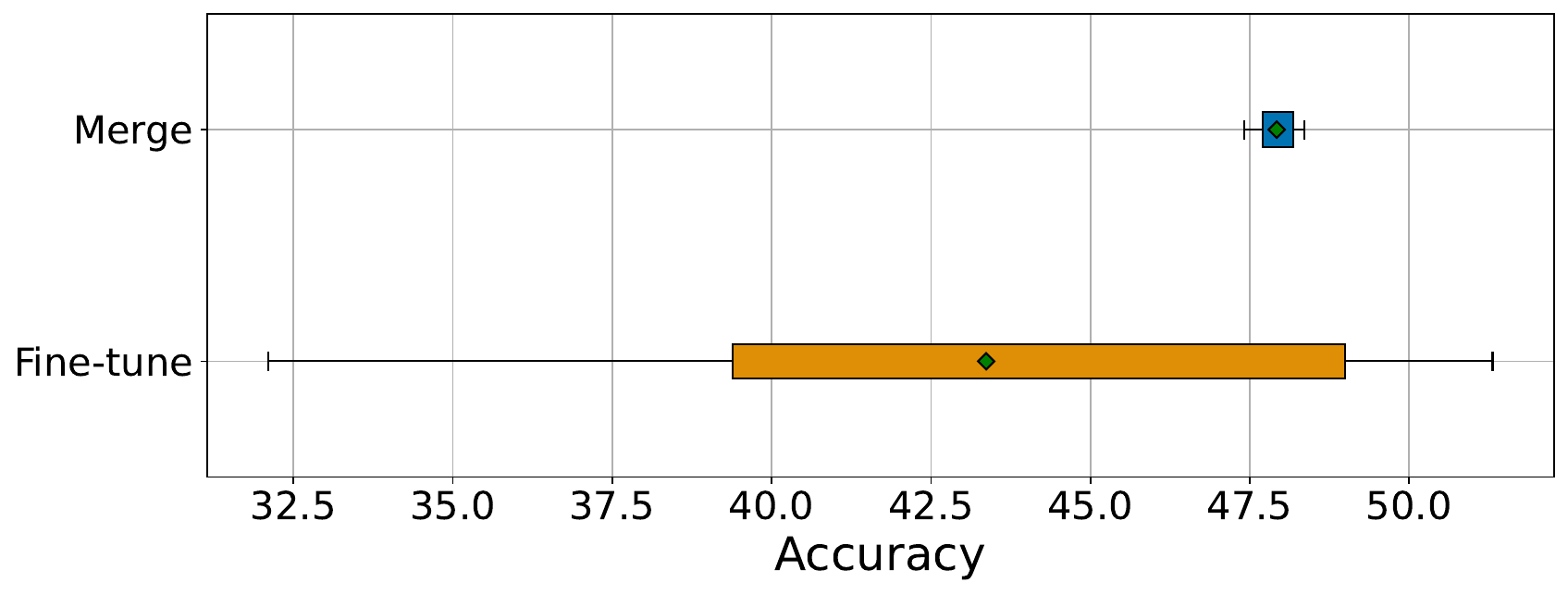}
\caption{(Left) Forgetting over time throughout training on visual QA tasks from the CoIN benchmark \citep{chen2024coin}, measured as the negation of the backward transfer metric described in \cref{sec:experimental_details}. At each time step, we report the average forgetting on seen tasks (lower is better). Merging demonstrates greater robustness to forgetting and task similarity compared to fine-tuning a single LoRA module, where performance fluctuates between steps. (Right) Average accuracy across all tasks for three different task orders. Fine-tuning shows higher sensitivity to task order.}
\label{figure1}
\end{center}
\vskip -0.2in
\end{figure}

We argue that catastrophic forgetting arises from the inherent bias of sequential fine-tuning towards the recently learned task, which favors plasticity over stability. We demonstrate that addressing the continual learning problem through the lens of model merging can effectively address the stability-plasticity dilemma. Specifically, after learning a new task, the new knowledge is integrated with the prior one by merging the latest learned weights with the weights of a global, continuously evolving model. We learn each task using LoRA modules for its computational efficiency. We show that this simple change in the learning paradigm offers various advantages, addressing the shortcomings of conventional fine-tuning such as catastrophic forgetting, sensitivity to task order and similarity, and lower generalization (see \cref{figure1,sec:merging} for full details). Furthermore, merging the weights in one global model addresses the parameter growth of previous techniques \citep{wang2023orthogonal}. 

To further improve knowledge integration across tasks without interference, we propose PAM, a simple \textit{during-training} {\bf P}arameter {\bf A}lignment method for effective {\bf M}erging. 
Specifically, while training a new task, we encourage alignment between the LoRA weights being learned and the weights of a global, continuously evolving LoRA module representing previous tasks (see schematic illustration in \cref{fig:schematic}).
After training, the weights of the current task are merged into the global LoRA. We demonstrate that this simple change in the conventional learning paradigm yields high robustness against forgetting over standard sequential fine-tuning. 

Our main contributions are:    
\begin{itemize}
    \item We demonstrate that leveraging model merging for continual learning, rather than the typical sequential fine-tuning, offers a solution to the latter's inherent shortcomings, such as its bias towards recent tasks, sensitivity to task orders and similarities, and poorer generalization. 
    \item We propose PAM (Parameter Alignment for Merging), a simple yet effective and computationally efficient approach that optimizes parameter alignment during task training to reduce inter-task interference and enable effective merging.
    \item We empirically demonstrate the effectiveness of PAM in improving performance on seen tasks and generalization to unseen ones across a diverse set of continual learning benchmarks. Furthermore, we show that it can be combined with existing continual learning techniques, bringing a further boost in performance. 
\end{itemize}
\section{Problem formulation}
\newcommand{\loss}{\mathcal{L}}
\newcommand{\Dist}{\mathcal{D}}
\newcommand{\EE}{\mathbb{E}}
We are presented with a sequence of tasks $[T]=\{1,\dots,T\}$, where for each task $t$, we are given i.i.d.~data $D_t$ sampled from an unknown distribution $\Dist_t$. Our aim is to find a model $f(\cdot)$ which minimizes the expected loss $\loss$ averaged over all the tasks seen thus far:
\begin{equation}
\label{eq:objective}
\textstyle{
\frac{1}{t} \sum_{t'=1}^t  \EE_{(x,y)\sim \Dist_{t'}}\loss (f(x),y).}
\end{equation}

We consider models $f$ with pretrained weights $W_0$ that are fixed, and a low rank adapter, LoRA, with weights $W$ that are trained on the continual learning tasks. Specifically, let $W_0^l \in \mathbb{R}^{n \times m}$ be the pre-trained weights at layer $l$. Additional trainable parameters $W^l$ are added to the layer, where $W^l$ is composed of low rank matrices $A\in \mathbb{R}^{n \times r}$ and $B\in \mathbb{R}^{r \times m}$, and $r$ is the LoRA rank. This rank is typically smaller than the original matrix dimensions (i.e., $r \ll \min(n,m)$). The forward pass of the data can then be written as
\begin{equation}
\label{eq:lora}
W_0^l x + W^l x = W_0^l x + BAx, 
\end{equation}
where $x$ is the input of the layer.

Although our models are aware when a new task arrives,
we assume that (1) the data of previous tasks $\{D_1,\dots,D_{t-1}\}$ is not available when learning a new task $t$; and (2) the task identity of an input is not available at inference time. These constraints are generally reflective of the types of challenges real-world situations would induce and further complicate the direct optimization of \cref{eq:objective}, as we are unable to directly sample from, or iterate through, past tasks.

\section{What can merging offer over fine-tuning?}
\label{sec:merging}

The fundamental approach in continual learning involves continuous fine-tuning as new tasks arrive. On top of this core fine-tuning, CL methods integrate regularization or replay to improve stability \citep{buzzega2020dark,kirkpatrick2017overcoming,wang2024comprehensive}. Sequential fine-tuning, however, suffers from catastrophic forgetting: it offers a relatively low expected loss on the most recent task $t$, but the expected loss of previous tasks is likely to increase. In this work, we provide another paradigm to address this challenge via merging, demonstrating its inherit ability to address the stability-plasticity dilemma. Just as fine-tuning, our approach can be combined with other continual learning approaches to further boost performance.

The modification to the conventional learning paradigm is rather simple: fine-tuning a temporary LoRA module on a new task, and then merging the learned parameters into a separate global LoRA module that evolves over time. Specifically, when a new task $t$ arrives, the current global LoRA weights, denoted $W_{G,t-1}$, are cloned into a separate LoRA module (with parameters $W'$) and fine-tuned on the new task data, $D_t$. After training on task $t$, the updated global LoRA weights, $W_{G,t}$, are obtained by merging $W_{G,t-1}$ and $W'$ using element-wise averaging, a standard model merging technique \citep{wortsman2022model,choshen2022fusing}.
This approach allows the model to integrate parameters learned from previous tasks while adapting to new tasks, effectively mitigating catastrophic forgetting in a parameter-efficient manner.

\cref{figure1} demonstrates the impact of this approach on a recent continual instruction tuning benchmark with visual question answering (QA) tasks \citep{chen2024coin} (Details are provided in \cref{sec:experimental_details}). We find that merging alone \textit{effectively mitigates catastrophic forgetting} of previous tasks, even without incorporating sophisticated mechanisms. We also observe that in fine-tuning a single LoRA module, performance of some tasks can fluctuate depending on the most recently learned task, as demonstrated by the increases and decreases in forgetting over time. In contrast, merging is \textit{less impacted by varying degrees of task similarity}. Notably, we find that merging exhibits \textit{more robustness to task ordering}.
This robustness to task ordering is a critical advantage over traditional fine-tuning, which, as shown in \cref{figure1} (right), is highly sensitive to the sequence of tasks.

\begin{figure*}
\vskip 0.2in
\begin{center}
\centerline{\includegraphics[width=0.92\textwidth]{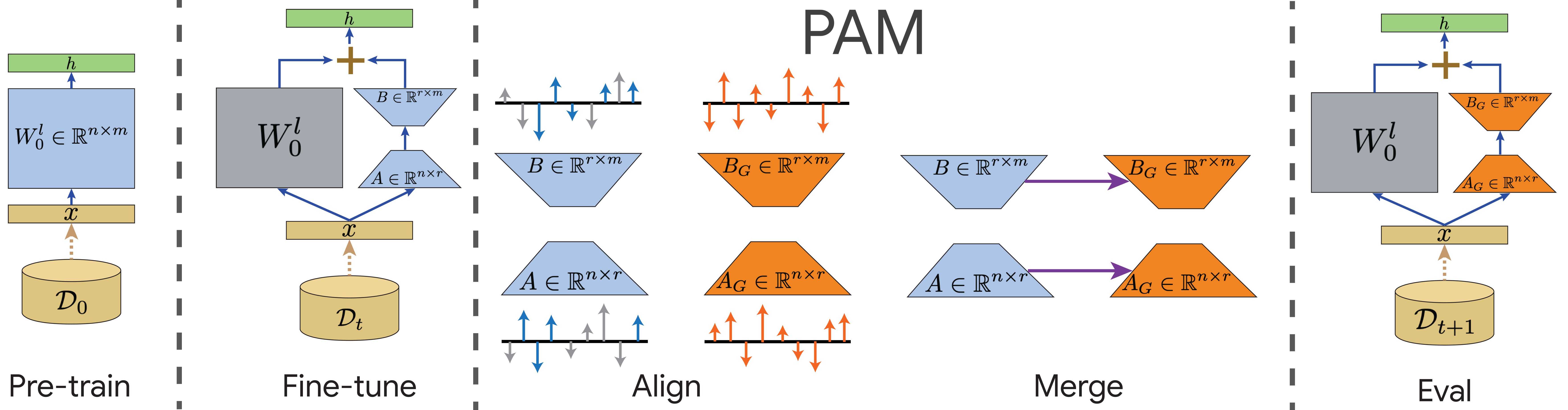}}
\caption{Method overview. \textcolor{gray}{After pre-training $W^l_0$}, PAM fine-tunes \textcolor{RoyalBlue}{temporary LoRA weights $\{A, B\}$} on a new task, and periodically reinitializes a subset of the parameters which are misaligned with \textcolor{orange}{a global LoRA $\{A_G, B_G\}$}. After training, the temporary LoRA $\{A, B\}$ is merged with the global LoRA $\{A_G, B_G\}$, which is ultimately used for evaluation.}
\label{fig:schematic}
\end{center}
\vskip -0.2in
\end{figure*}

\section{Parameter Alignment and Merging (PAM)}
\label{sec:method}

\begin{wrapfigure}{r}{0.45\textwidth}
\begin{minipage}{\linewidth}
\begin{algorithm}[H]
   \caption{PAM}
   \label{alg:palm}
\begin{algorithmic}
   \STATE {\bfseries Input:} pretrained weights $W_{0}$, Global LoRA $W_{G,t-1}$, alignment percent $p$, schedule $s$, number of tasks $T$
   \FOR{$t=2$ {\bfseries to} $T$}
   \STATE Add a new LoRA with parameters $W'$
   \STATE $W'$ = $W_{G,t-1}$
    \STATE A = Sort $|W_{G,t-1}|$ in descending order
    \STATE $k = \lceil (p / 100) \cdot \text{length}(A) \rceil$ 
    \STATE $ \tau = A[k]$ 
   \FOR{$step=1$ {\bfseries to} $total\_steps$}
   \STATE Update $W'$ 
   \IF{$step \mod s == 0$}
   \STATE // Check misalignment 
   \FOR{each weight $w$ in $W'$}
\IF{$\text{sign}(w') \neq \text{sign}(w_{G,t-1}) \, \wedge \, w_{G,t-1} > \tau$}

   \STATE Re-initialize $w'$ 
   \ENDIF
   \ENDFOR
   \ENDIF
   \ENDFOR
   \STATE // Merge
   \STATE $W_{G,t}$ = Avg($W_{G,t-1},W'$)
   \ENDFOR
\end{algorithmic}
\end{algorithm}
\end{minipage}
\vspace{-2em}
\end{wrapfigure}
\cref{sec:merging} demonstrates the effectiveness of model merging in addressing the CL challenges with a simple 
post-training paradigm shift. However, while direct merging of weights learned from different tasks shows promise compared to standard fine-tuning, simply averaging these weights can be suboptimal as task-specific updates may interfere with each other. 
To address this, we propose to optimize for Parameter Alignment \textit{during} training for effective Merging, leading to our method PAM.

PAM introduces a simple change to new task training to encourage alignment with prior tasks, as depicted in \cref{fig:schematic}. 
\textit{During} new task training, we \textit{periodically} check the current task weights, identifying and re-initializing any that become misaligned with the important weights of the global LoRA weights, that capture knowledge of prior tasks. To measure alignment, we leverage previously published findings, demonstrating that the weight sign is a simple and effective tool for alignment in both weight and gradient spaces \citep{chen2020just, yadav2024ties, davari2025model}.

Specifically, during task training, we periodically compare the signs of $W'$ with those in $W_{G,t-1}$, and re-initialize the weights in $W'$ that have misaligned signs, provided their corresponding weights in $W_{G,t-1}$ are among the top $p$\% in magnitude. The alignment percentage $p$ acts as a control parameter, balancing the model's stability and plasticity. A higher $p$ emphasizes preserving previously learned knowledge (stability) at the potential cost of reduced adaptability to the current task (plasticity) (see \cref{sec:effect_of_alignment}). After training on task $t$, the updated global LoRA weights, $W_{G,t}$, are obtained by merging $W_{G,t-1}$ and $W'$. We provide an outline in \cref{alg:palm}.  

Although many techniques have been recently proposed for improving model merging
\citep{yadav2024ties,wanglocalizing}, they operate \textit{post-training}, in contrast to our method that operates \emph{during training}. For instance, a closely related method, TIES \citep{yadav2024ties}, utilizes post-training alignment by \textit{pruning} weights whose signs conflict with the sign of the highest total magnitude across all LoRAs. We argue that periodic alignment during training offers several advantages. It prevents LoRA weights from diverging excessively during fine-tuning, maintaining lower sign conflicts throughout training; moreover, periodic alignment allows weights to recover from initial misalignment and continue contributing to the learning process, avoiding the capacity loss associated with pruning in TIES, as discussed further in \cref{sec:effect_of_alignment}. This dynamic alignment strategy enables better knowledge integration and effective merging compared to a one-time post-training alignment.

\section{Experiments}
\subsection{Experimental Details}
\label{sec:experimental_details}
\paragraph{Benchmarks} We evaluate our method on CoIN \citep{chen2024coin}, a recent CL benchmark for multimodal large language models. We considered the six image QA tasks: VQAV2 \citep{goyal2017making}, TextVQA \citep{singh2019towards}, OCR-VQA \citep{mishra2019ocr}, ScienceQA \citep{lu2022learn}, VizWizVQA \citep{gurari2018vizwiz}, and GQA \citep{hudson2019gqa}. For this benchmark, we evaluate three different task orderings: the original order and the alphabetically sorted order introduced in \citep{chen2024coin}, where the former proved to be more challenging, and a randomly generated order. While we provide the main results for all three, our extended analysis is centered on the most challenging original task order. Furthermore, we extend CoIN by adding four image-based tasks—AI2D \citep{kembhavi2016diagram}, AOKVQA-MC \cite{schwenk2022okvqa}, OKVQA \citep{marino2019ok}, and RSVQA-lr \citep{lobry2020rsvqa}—resulting in a longer sequence of ten tasks with higher diversity and task dissimilarity, including remote sensing VQA and multi-choice QA. Finally, we evaluate on a sequence of video QA that includes MSVD-QA \citep{xu2017video}, MSRVTT-QA \citep{xu2017video}, and ActivityNet-QA \citep{yu2019activitynet}, further diversifying the evaluation beyond image-based tasks. 
See \cref{appendix:experimental_details} for further details.

\paragraph{Model, code, and compute resources} As our base model we use PaliGemma \citep{beyer2024paligemma}, a representative and open-source VLM known for its effectiveness across a wide range of transfer tasks. Inputs are set to a resolution of $224 \times 224$ pixels, and we apply the best hyperparameters reported in \citep{beyer2024paligemma} for each task, except for the learning rate, which we adjust to $5 \times 10^{-4}$ for improved performance with LoRA training. We use a LoRA rank of $r=32$ and alignment percentage $p= 50\%$. Hyper-parameter details are presented in \cref{appendix:experimental_details}. For all our experiments, we build on top of the big vision library 
\citep{big_vision}, a Jax and Flax-based framework. The library and the benchmark used have Apache2 license. All experiments were run on TPUv2, with each individual task in the sequence utilizing 64 TPUs. The duration of training each task ranges from 1 hour to 10 hours depending on the size of the task, with a six-task sequence completing in roughly 20 hours.

\paragraph{Baselines} We compare against several state-of-the-art methods. We start with standard approaches: (1) \textit{Zero-shot}: tasks are evaluated using the pre-trained PaliGemma model without further training; (2) \textit{Independent}: Separate LoRA modules are fine-tuned on each task independently; (3) \textit{Multitask}: A single LoRA is trained concurrently on all tasks; (4) \textit{Fine-tuning}: A single LoRA is trained sequentially on all tasks. We also compare against continual learning methods including: (5) LWF \citep{li2017learning}: a well-known regularization method that adds a distillation loss to restrict the change in previous knowledge; (6) I-LoRA \citep{li2025analyzing}: the framework consists of two LoRA modules (faster LoRA learner and slower LoRA learner), and the knowledge is slowly consolidated during training in the slower learner. Note that this method initially incorporates experience replay of previous task data, but we exclude this part for fair and direct comparison against all rehearsal-free baselines; (7) O-LoRA \citep{wang2023orthogonal}: it adds new LoRA module for each new task, while optimizing its weights to be orthogonal to previously added LoRA modules to reduce interference. (8) MoELoRA \citep{chen2024coin}: a mixture of experts is trained on the CL sequence with learned gate function to route the input to experts; (9) MaxMag \citep{marczak2024magmax}: a closely related work that merges knowledge from all previous learned models by picking the parameter with the highest magnitude; we applied this technique to LoRA weights. Finally, we compare against state-of-the-art model merging techniques proposed for multitask learning including (10) \textit{Average}: LoRA weights are merged using a simple average; (11) \textit{TIES} \citep{yadav2024ties}: this post-training technique reduces interference by \textit{pruning} the weights that are misaligned; (12) \textit{TALL} \citep{wanglocalizing}: this technique employs a post-training localization strategy to identify and merge important weights and can complement other merging techniques that employs parameter alignment. For the merging baselines, we maintain one core LoRA, which is copied as a new task arrives, fine-tuned on that task, and then merged back with the core. Note that these methods were initially designed with independent initialization for each task. However, we discovered that using the initialization of the core module as a starting point yields better results, as we demonstrate in \cref{sec:lora_initalization}. Therefore, we will compare against the improved version of these baselines.  
    
\paragraph{Evaluation metrics} We do extensive evaluation by assessing four metrics to measure  model capabilities: (1) \textit{Accuracy (ACC)}: the average accuracy of all tasks at the end of the sequence of $T$ tasks; (2) \textit{Backward transfer} (BWT) \citep{lopez2017gradient}: this metric measures the influence of learning task $t$ on the performance of each previous task $i < t$. Thus, it assess the model robustness against forgetting; (3) \textit{Forward transfer} (FWT) \citep{lopez2017gradient}: this metric measures the ability of a model to generalize to unseen tasks. It is estimated by the difference between the model performance at time step $t$ on unseen tasks ($i > t$) and the zero-shot performance. (4) \textit{Average task Accuracy} ($A_t$): this metric assess the plasticity of the model to learn a new task. It measures the performance of task $t$ immediately after learning it. For merging based techniques, we also report $A_m$, which represents the accuracy after the merging step. The computation details of each metric can be found in the appendix. 

\subsection{Results}
Motivated by recent observations of strong zero- and few-shot learning performance in large models \citep{brown2020language}, we start by evaluating the performance of the pretrained PaliGemma model on CoIN tasks. Despite its large pretraining data, we observe that its generalization to these downstream tasks is suboptimal, as evidenced by the zero-shot performance when compared to adapting an independent LoRA for each task or training single LoRA for all tasks jointly as shown in \cref{table:coin_results}. Although sequential fine-tuning improves performance and has high plasticity ($A_t$), it suffers from catastrophic forgetting and poor generalization, as reflected by the lower BWT and FWT scores.   

\begin{table*}[t]
\caption{Performance of different methods on the CoIN benchmark across four metrics that assess various model capabilities. Results are presented as the mean and standard deviation over three task orders, with higher values indicating better performance for all metrics.}
\label{table:coin_results}
\vskip 0.15in
\begin{center}
\begin{sc}
\resizebox{0.7\textwidth}{!}{%
\begin{tabular}{lcccc}
\toprule
Method & ACC & BWT & FWT & $A_{t}$\\
\midrule
Zero-shot  &  24.74 & - & - & -\\
Independent &76.46  & -& - & - \\ 
Multitask &  73.93 & - & - & -\\
\hline
Fine-tune &  43.36$\pm$8.18&-39.51$\pm$9.87& 7.71$\pm$2.51&  76.29$\pm$0.18\\
LWF & 47.15$\pm$3.52 & -33.66$\pm$3.84  & 9.67$\pm$1.22 & 75.20$\pm$0.32\\
I-LoRA & 42.11$\pm$6.55 & -40.95$\pm$8.13 & 7.89$\pm$2.60 & 76.24$\pm$0.35\\
O-LoRA & 46.53$\pm$6.88 & -32.08$\pm$7.87 & 9.45$\pm$3.02 & 73.27$\pm$1.01  \\
MoELoRA & 46.59$\pm$9.98 & -36.40$\pm$11.97 & 7.79$\pm$2.24 & \textbf{76.93$\pm$0.27}\\
MagMax &45.74$\pm$0.88 & -22.68$\pm$6.51 & 4.75$\pm$3.36&  76.29$\pm$0.18\\ 
PAM (ours) & \textbf{49.89$\pm$1.66} &  \textbf{-19.45$\pm$0.95} & \textbf{11.11$\pm$0.09} & 76.31$\pm$0.03\\   

\bottomrule
\end{tabular}
}
\end{sc}
\end{center}
\vskip -0.1in
\end{table*}

Merging aligned models via PAM demonstrates significant performance improvements over standard fine-tuning, exhibiting enhanced robustness against forgetting even without explicit forgetting mitigation techniques. Furthermore, PAM enhances the model's generalization capabilities and increases forward transfer across tasks. Interestingly, PAM surpasses methods that expand parameters over time (O-LoRA), or utilize a mixture of experts (MoELoRA), despite relying solely on a single evolving LoRA for all tasks. Moreover, we find that the averaging of aligned weights in PAM is more effective than selecting the parameter with the highest magnitude from each learned LoRA as in MagMax.  

\citet{chen2024coin} observed that task order influences model performance. This is likely attributable to the impact of the sequence on the learning dynamics and trajectory, which are shaped by task complexity, data size, and dissimilarity. We find that fine-tuning a single or mixture of LoRAs exhibits greater sensitivity to task order, as demonstrated by the high standard deviation in performance across different sequences. This is also highlighted by the fluctuations in task performance at each step during fine-tuning, as illustrated in \cref{figure1} and  \cref{appendix:extra_results}. In contrast, merging-based methods, specifically MagMax and PAM, demonstrate greater robustness to task order, delivering more stable performance regardless of the task sequence. 

\subsection{Empirical Analysis}
\label{sec:effect_of_alignment}
In this section we analyze the effect of \textit{during-training} alignment, which is the key algorithmic advancement behind our proposed method, PAM. 
Specifically, we present ablations investigating how during-training alignment affects PAM, revealing the role of this step in mitigating catastrophic forgetting. We also compare PAM against post-training interventions for improved merging, and examine the impact of PAM's design choices.

\begin{table*}[t]
\caption{Comparison of various merging methods on the CoIN benchmark. Results are presented as the mean and standard deviation over three task orders.}
\label{table:alignment_during_vs_post}
\begin{center}
\begin{sc}
\resizebox{0.9\textwidth}{!}{%
\begin{tabular}{lcccccc}
\toprule
Method & Alignment/Localization& ACC & BWT & FWT & $A_{t}$ &$A_{m}$\\
\midrule
Average & $\times$ & 47.41$\pm$0.38 & -22.40$\pm$0.80& 10.50$\pm$2.24 & 76.31$\pm$0.15 &65.95$\pm$1.31 \\
PAM (ours) & during training & 49.89$\pm$1.66 &  -19.45$\pm$0.95 & 11.11$\pm$0.09 & 76.31$\pm$0.03 &66.11$\pm$0.87   \\

TIES    & post training  & 47.78$\pm$0.38 & -22.87$\pm$0.68& 9.92$\pm$2.05 & 76.35$\pm$0.14 &66.85$\pm$0.26 \\
TALL & post training & 49.22$\pm$2.53 & -12.85$\pm$1.53 & 11.90$\pm$0.57 & 76.42$\pm$0.12 &59.93$\pm$1.51     \\

TALL+PAM & During \& Post training &  51.97$\pm$5.59 &  -9.27$\pm$5.53 & 11.96$\pm$0.97 &76.28$\pm$0.25  &59.70$\pm$1.23 \\
\bottomrule
\end{tabular}
}
\end{sc}
\end{center}
\vskip -0.1in
\end{table*}

\begin{figure}[t]
  \centering
    \begin{minipage}{0.33\linewidth}
        \centering
\includegraphics[width=\columnwidth]{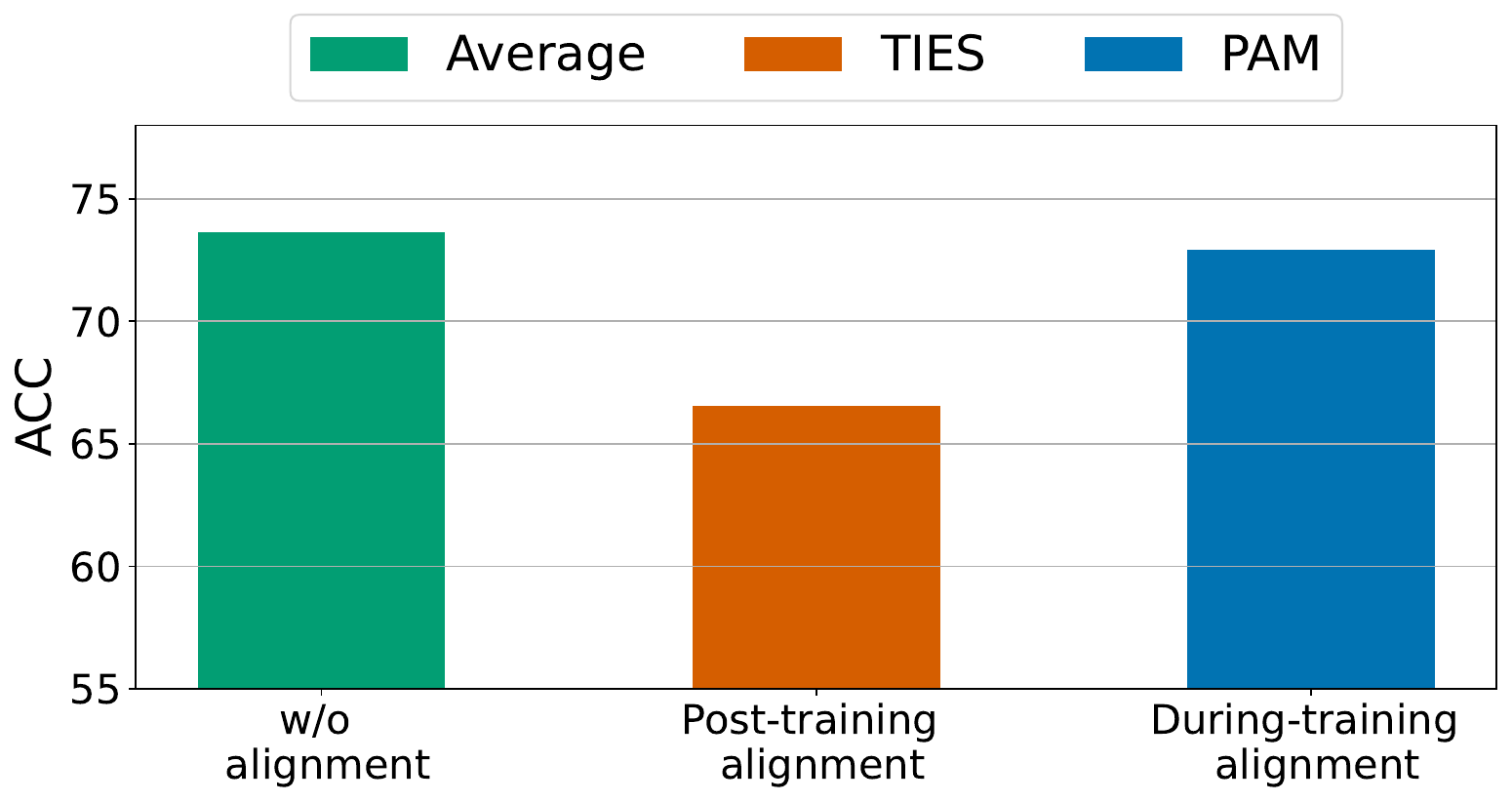}
\caption{Task accuracy before merging. Pruning misaligned parameters after training decreases performance, whereas optimizing for alignment during training maintains model plasticity.}
\label{fig:ties_vs_palm}
\end{minipage}
  \hspace{0.3cm}
  \begin{minipage}{0.6\linewidth}
    \centering
\includegraphics[width=0.49\columnwidth]{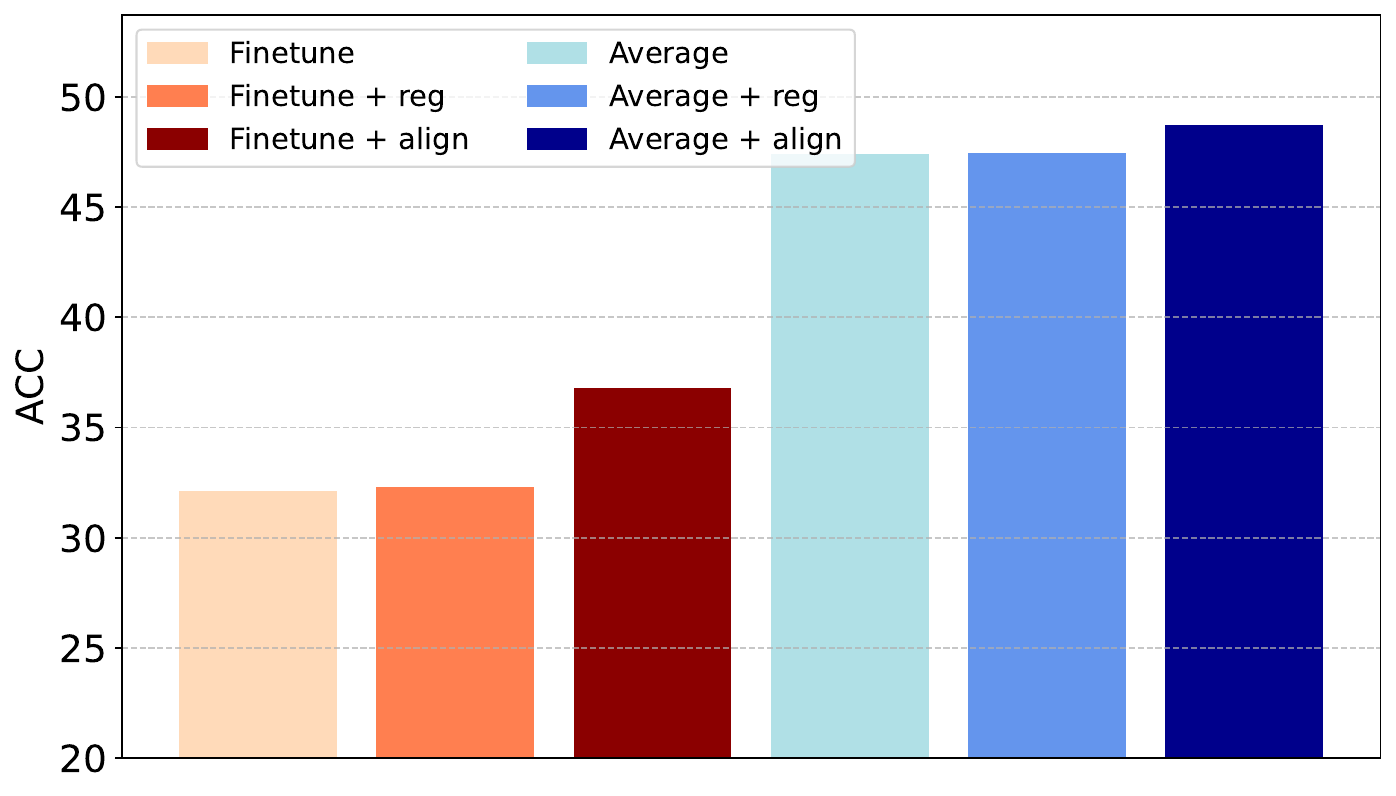}
\includegraphics[width=0.49\columnwidth]{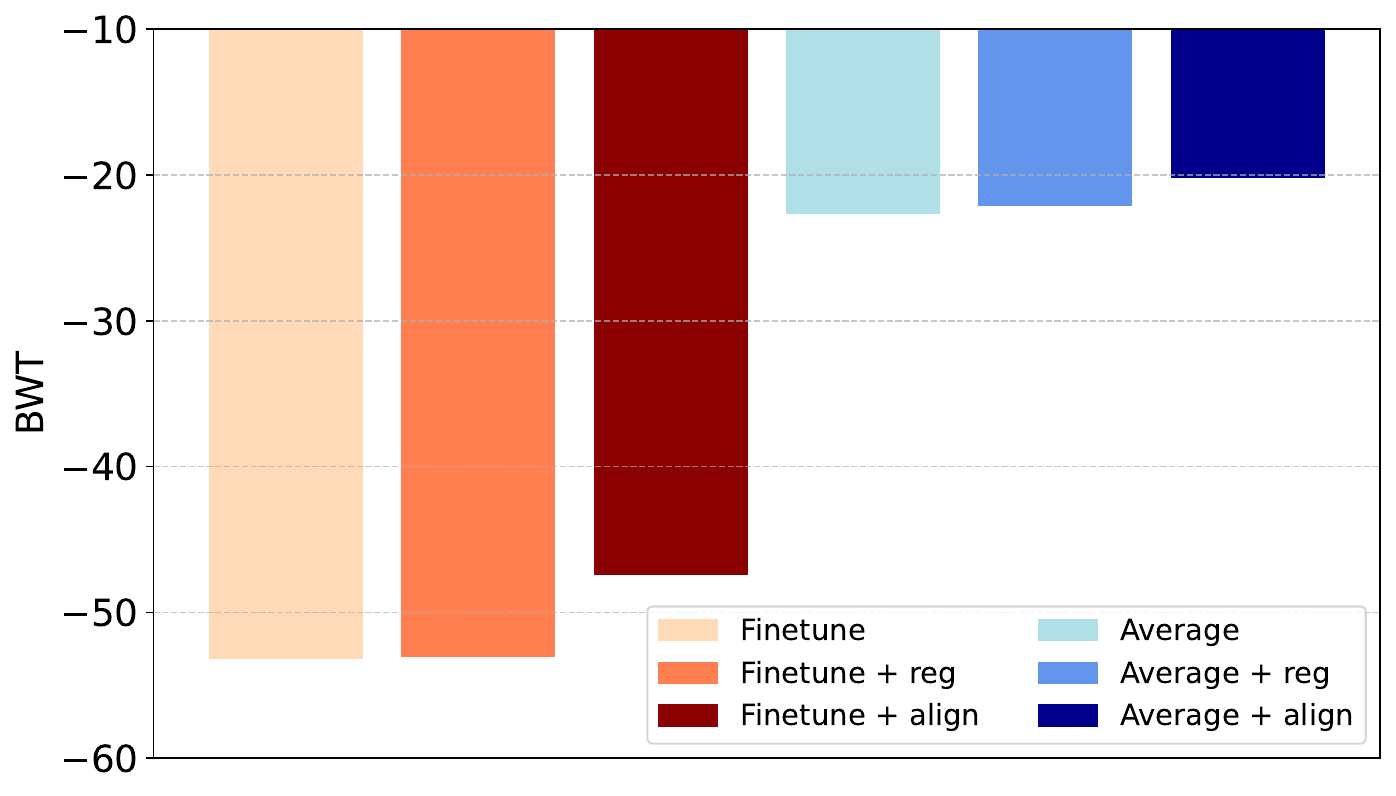}
    \caption{Weight regularization provides only limited benefits in mitigating forgetting. In contrast, alignment improves performance of fine-tuning and average baselines.}
    \label{fig:regvsalign}
   \end{minipage}
   
\end{figure}

\paragraph{Improved knowledge retention and reduced task interference} \cref{table:alignment_during_vs_post} presents a comparison against state-of-the-art merging methods. The high performance of different strategies further confirms the gains by merging over the conventional fine-tuning. To assess the effect of during-training alignment in isolation, we compare against the average baseline, which is equivalent to PAM but with the alignment process ablated. As shown in the table PAM has superior performance over this baseline across all metrics. Comparing PAM and TIES, both utilizing parameter alignment, shows that aligning parameters during training improves knowledge retention and reduces task interference. Post-training alignment often favors one task when misalignment occurs, and pruning misaligned parameters can harm accuracy, as we will demonstrate next. In contrast, during-training alignment encourages the model to find parameters that aligned with previous tasks, leading to more effective merging of knowledge after training. Localizing and merging important weights between a global LoRA and a temporary one by TALL yields good average accuracy, but low $A_m$. This could be due to pruning selfish weights by TALL that are important to one task. Integrating PAM's during-training alignment with TALL's post-training localization effectively improves TALL's performance and reduces forgetting, further demonstrating the effectiveness of during-training alignment.

\paragraph{Improved model plasticity and capacity} We examine the impact of alignment on task accuracy and analyze the proportion of misaligned parameters. In this analysis, we focus on the final two tasks of one sequence, which demonstrates some level of dissimilarity. We compare the behavior of post-training alignment (TIES) with that of during-training alignment (PAM). Our findings reveal that, on average, 29.41\% of parameters across all layers are misaligned between the global module and the most recent LoRA module trained using TIES. As illustrated in \cref{fig:ties_vs_palm}, just applying post-training alignment to the penultimate task prior to merging results in a 7.44\% reduction in accuracy. In contrast, during-training alignment preserves model plasticity, as evidenced by PAM's accuracy, which remains comparable to the case of without alignment. 


\begin{figure}[b]
  \centering
    \begin{minipage}{0.45\linewidth}
        \centering
\resizebox{\textwidth}{!}{%
\begin{tabular}{lccccc}
\toprule
Alignment [\%] & ACC & BWT & FWT & $A_t$&$A_m$ \\
\midrule
30\% & 47.74 & -22.24 & 11.21 &76.49 &66.28\\
50\% & 48.16 & -21.37 & 11.04 & 76.33&65.97\\
70\% & 48.32 & -20.63 & 11.19 & 76.30&65.51 \\
\bottomrule
\end{tabular}
}
\captionsetup{type=table}
\caption{Effect of the percentage of alignment in PAM.}
\label{table:effect_align_percentage}
\end{minipage}
  \hspace{1cm}
  \begin{minipage}{0.45\linewidth}
    \centering
\resizebox{\textwidth}{!}{%
\begin{tabular}{lccccc}
\toprule
Method & ACC & BWT & FWT & $A_t$ & $A_m$\\
\midrule
reinit $W_{G,t-1}$ & 49.89$\pm$1.66 &  -19.45$\pm$0.95 & 11.11$\pm$0.09 & 76.31$\pm$0.03 & 66.11$\pm$0.87\\
reinit zero & 49.53$\pm$1.61& -20.14$\pm$1.19& 10.81$\pm$0.34 & 76.24$\pm$0.15 &66.32$\pm$0.65\\
\bottomrule
\end{tabular}
}
\captionsetup{type=table}
\caption{Effect of the reinitialization strategy in PAM.}
\label{table:reinit_strategy}
   \end{minipage}
\end{figure}  

\paragraph{Regularization does not have the same effect} 
Regularization is a common CL strategy to preserve knowledge from past tasks by restricting the weights from moving too far from their original values \citep{kirkpatrick2017overcoming, aljundi2018memory}. We investigate whether regularizing the weights would be sufficient to promote parameter alignment and mitigate forgetting. 

Our results in \cref{fig:regvsalign} demonstrate that adding regularization yields a slight performance improvement, but it fails to effectively address catastrophic forgetting. Interestingly, we find that alignment can even \textit{provide gains to the fine-tuning baseline and reduce forgetting.} This further demonstrates the effectiveness of alignment and its potential for improving performance of other techniques.

\begin{figure}[t]
\begin{center}
\includegraphics[width=0.47\columnwidth]{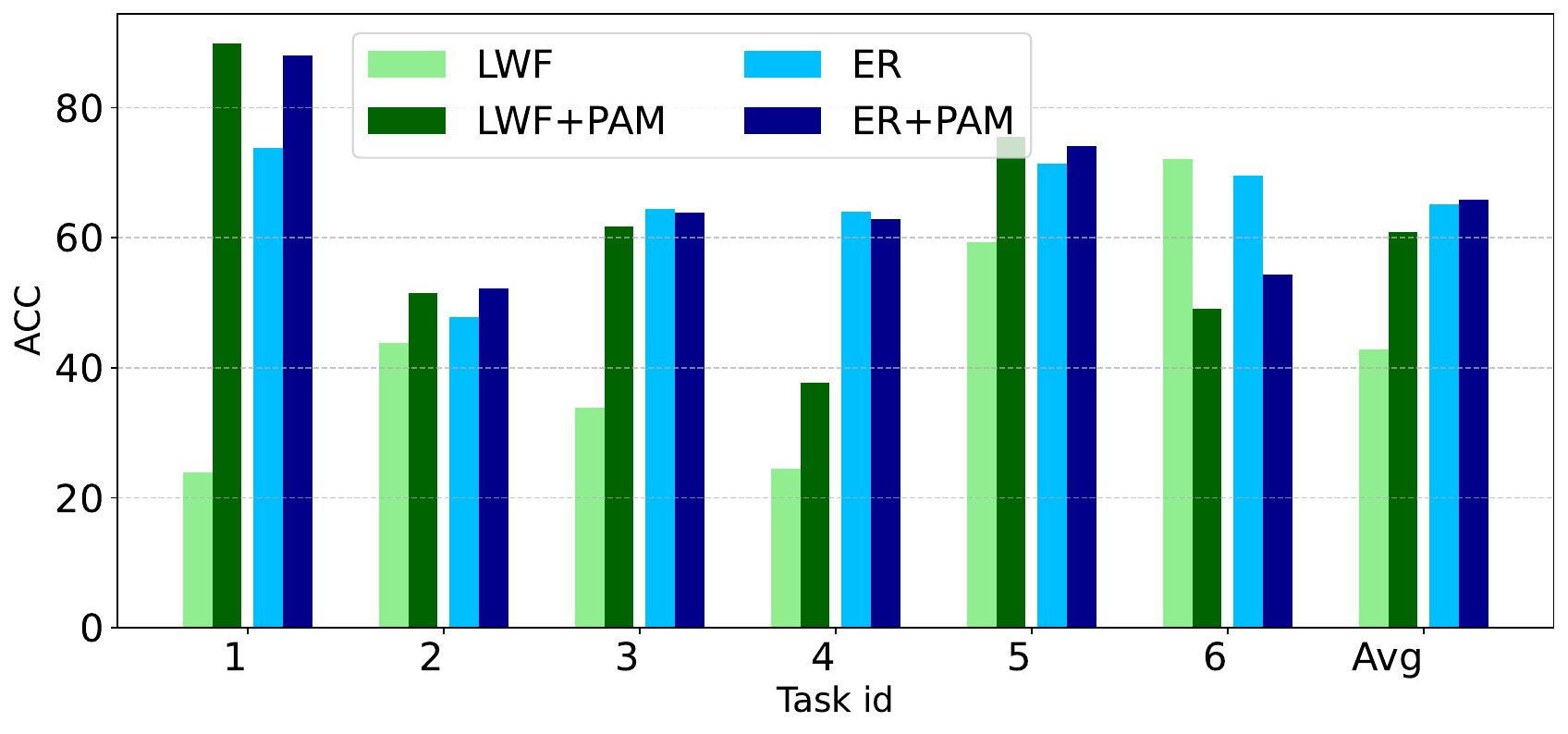}
\includegraphics[width=0.47\columnwidth]{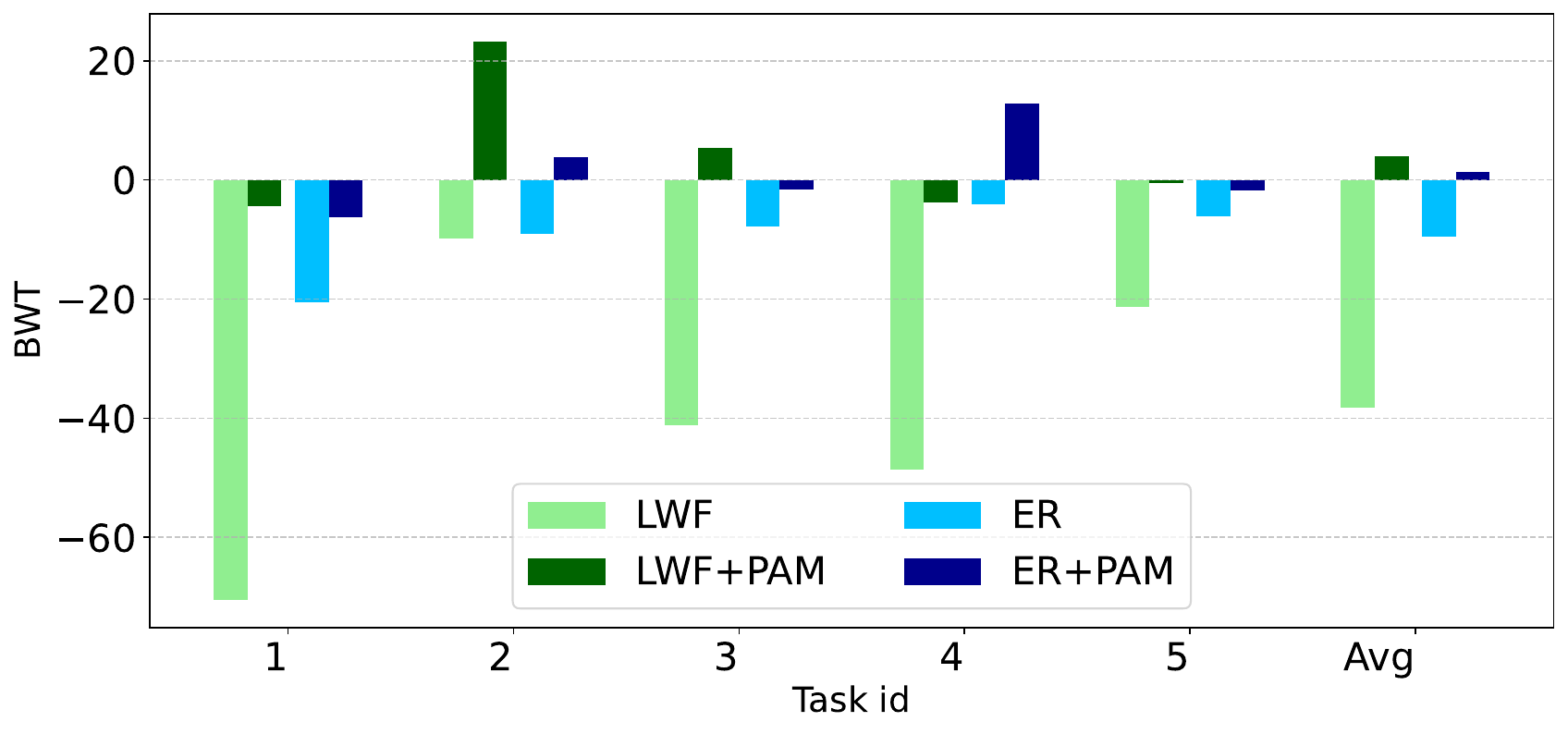}
\caption{Evaluating the performance of existing CL methods on the original order of the CoIN benchmark when combined with PAM. We report the ACC (left) and BWT (right) for each task after training on the whole sequence. PAM yields improved performance and less forgetting, particularly demonstrating its efficacy in challenging scenarios where past data is not available.}
\label{fig:replay_vs_palm_and_replay}
\end{center}
\vskip -0.2in
\end{figure}

\paragraph{Control over the stability-plasticity trade-off} 
During-training alignment, as opposed to post-training, allows controlling the percentage of aligned parameters. To investigate this effect, we analyze PAM with varying percentages of aligned parameters (30\%, 50\%, and 70\%), and report the results in \cref{table:effect_align_percentage}. We find that larger percentage of alignment favors previous tasks and reduces forgetting. 
Nevertheless, increasing the alignment percentage reduces the model's plasticity as shown by the reduced $A_t$ and $A_m$. 
This highlights the trade-off between stability and plasticity in CL and that PAM provides a valuable tool for navigating this trade-off.

\paragraph{Reinitialization strategy plays a role} We examine two strategies for reinitializing misaligned weights: (1) zero initialization, and (2) using global weights $W_{G,t-1}$. Our findings indicate that the two strategies are effective, with re-initializing with global weights favors stability and reduces forgetting at a slight cost to current task performance after merging, as shown in \cref{table:reinit_strategy}.


\subsection{PAM as a Complementary Continual Learning Technique}
\label{sec:palm_replay}
The simplicity of PAM facilitates its integration with existing CL techniques. In this section, we evaluate its performance when integrated with two representative methods from different strategies: LWF \citep{li2017learning}, regularization-based, and Experience Replay (ER) \citep{robins1995catastrophic, chaudhry2019tiny}, a rehearsal based strategy.    

\cref{fig:replay_vs_palm_and_replay} presents the accuracy and backward transfer on each task after learning the whole sequence of the CoIN benchmark. We find that PAM substantially improves the performance of earlier tasks, and reduces forgetting when integrated with LWF. Moreover, despite the effectiveness of ER in preserving performance on earlier tasks through replying old samples, PAM still provides additional gains. Interestingly, the combination of PAM with the regularization-based method LWF achieves performance comparable to ER performance, notably without old data. These findings demonstrate the potential of our proposed shift in the learning paradigm for improved performance.      

\begin{table}[t]
\caption{Performance on the extended CoIN and video QA benchmarks.}
\label{table:video_results}
\begin{center}
\begin{sc}
\resizebox{0.7\columnwidth}{!}{%
\begin{tabular}{lcccc|cccc}
\toprule
 & \multicolumn{4}{c|}{Extended CoIN} & \multicolumn{4}{c}{Video QA} \\
Method & ACC & BWT & FWT & $A_t$ & ACC & BWT & FWT & $A_t$\\
\midrule
Zero-shot    & 23.08  & - & - & -  & 22.06 & - & - & - \\
Independent & 69.83 & - & -& -& 51.27 & - & - & -\\ 
\hdashline
Fine-tune & 49.73 & -22.19 & 2.87 & 69.71 & 40.73 & -16.66 & 10.61 & 51.84\\
PAM  (ours) & 50.43 & -14.10 &9.38 & 69.64 & 46.06 & -1.95 & 11.79 & 51.75 \\ 

LWF & 49.49 & -20.81 & 3.56 & 68.23 & 41.55 & -14.60 & 11.04  & 51.30 \\
LWF + PAM & 51.53 &  -5.08&  7.60 & 67.90 & 46.12 & -0.46 & 11.69 & 51.29\\

\bottomrule
\end{tabular}
}
\end{sc}
\end{center}
\vskip -0.1in
\end{table}

\paragraph{Evaluation on additional benchmarks} 
Finally, we assess the generalization of our findings on two other benchmarks: extended version of CoIN with longer sequence of diverse tasks and video question answering tasks. \cref{table:video_results} presents the results on the four studied metrics. Consistent with previous observations, PAM reduces forgetting and increase forward transfer. Moreover, it improves performance of LWF.

\section{Related Work}
\label{sec:relatedwork}
\paragraph{Model merging} 

Model merging has emerged as a promising technique for combining multiple models fine-tuned on different tasks.
Several methods have shown to perform well, with a few dominating the landscape, such as averaging \citep{izmailov2018averaging,wortsman2022model}, model arithmetic \citep{ilharco2022editing}, TIES \citep{yadav2024ties}, and TALL-masks \citep{wanglocalizing}.
These methods have been primarily applied to merging entire models, and the success of these merging techniques has been shown to rely on linear mode connectivity \citep{frankle2020linear,sharma2024non}. 
Recent concurrent work studied merging entire models over time \citep{kleiman2025soup,dziadzio2024merge}, with \cite{dziadzio2024merge} revealing that current merging methods offer little benefit over simple averaging for temporal merging.
More recently, merging fine-tuned LoRAs instead of full models has been explored. Initial studies suggest, however, that directly applying existing techniques to LoRA merging does not yield substantial gains. 
For instance, \citet{prabhakar2024lora} investigated different merging strategies applied to LoRAs and found that simple concatenation often outperforms more sophisticated methods in a skill composition task. This highlights the need for further research to develop specialized methods tailored to the unique characteristics of LoRAs.

\paragraph{Continual learning} The CL paradigm includes several key objectives, such as mitigating forgetting on past tasks, facilitating forward transfer to leverage prior knowledge for new tasks, and ensuring computational efficiency in learning, among others \citep{diaz2018don,hadsell2020embracing}. Most prior work focuses on addressing forgetting, which can be categorized into regularization, architectural, and replay-based methods. Regularization methods minimize forgetting by constraining changes to important weights \citep{aljundi2018memory,kirkpatrick2017overcoming,ahn2019uncertainty}. Architectural methods modify the model’s structure, often using separate modules for different tasks, to reduce interference \citep{wortsman2020supermasks, sokar2021spacenet,konishi2023parameter,gurbuz2022nispa}. Replay-based methods incorporate a subset of previous task data alongside new task data during training \citep{buzzega2020dark,rebuffi2017icarl,shin2017continual}. This paper introduces a perspective that is orthogonal to the three CL categories, focusing on model merging rather than standard fine-tuning which is the core component in these approaches. A few recent works explored adding new adapters over time \citep{liu2024learning,yu2024boosting,wang2023orthogonal}. \citet{biderman2024lora} showed that LoRA fine-tuning results in less forgetting of the base model's knowledge compared to full fine-tuning, having less impact on performance for tasks beyond the downstream task. \citet{marczak2024magmax} proposed integrating knowledge from each task model into one by selecting the parameters with the highest magnitudes across all models.  
\section{Discussion and Conclusion}
\label{sec:conclusion}
In this work, we demonstrated key limitations of fine-tuning within the CL paradigm, such as a strong bias toward recent tasks, reduced generalization capability, and increased sensitivity to task order. To address these challenges, we propose a novel view that replaces standard fine-tuning with a merging-based strategy to better integrate new and previously learned knowledge. Our proposed method, PAM, incorporates two key components. First, separate fine-tuning on the latest task followed by merging with previously learned weights. 
Second, to address potential interference between tasks, we introduce periodic alignment of the current task parameters with those from prior tasks during training. To achieve computationally efficient training on a continuous data stream, we build our approach on LoRA.

We conducted evaluations across various vision-language benchmarks, considering differences in sequence length, complexity, and task diversity. The results indicate that PAM mitigates forgetting and enhances generalization. Additionally, it demonstrates greater robustness to variations in task order. Furthermore, our findings reveal that alignment during training surpasses state-of-the-art merging techniques that rely on post-training alignment, and even provides gains to sequential fine-tuning without merging. Lastly, we showed that PAM improves existing CL approaches.

\paragraph{Limitations and future work} While the straightforward merging of data into a single LoRA offers an attractive solution for scaling in the CL paradigm, it may not adequately address scenarios where new data significantly differs from previously learned tasks. Additionally, selectively fine-tuning subsets of the weights could help mitigate forgetting and enhance the forward transfer of relevant knowledge. 

\section*{Acknowledgements}
The authors would like to thank Clare Lyle, Razvan Pascanu, Hugo Larochelle, Alexandre Ramé, Doina Precup, and the rest of the Google DeepMind team for valuable feedback on this work.

\bibliography{main}



\newpage
\appendix
\section{Experimental Details}
\label{appendix:experimental_details}
\begin{table}[t]
\caption{Hyperparameters used in training each task. We used the best hyperparameters reported by \citet{beyer2024paligemma} on fine-tuning on each task independently.}
\label{table:hyperparameters}
\vskip 0.15in
\begin{center}
\begin{small}
\begin{sc}
\begin{tabular}{lcccc}
\toprule
Task & Epochs & Batch size & Weight Decay & Label Smoothing \\
\midrule
ScienceQA & 20 & 128 & 0.0 & 0.0\\
TextVQA & 5 & 256 & 0.0 & 0.0\\
GQA &  1 & 256 & 0.0 & 0.0 \\
VizWizVQA & 10 & 256 & 0.0 & 0.0\\ 
VQAv2 & 10 & 256 & 1e-06 & 0.0  \\
OCR-VQA & 3 & 128 & 0.0 & 0.0  \\
RSVQA-lr & 3 & 256 & 0.0 & 0.2 \\
AI2D & 10 & 256 & 1e-06 & 0.0 \\
OKVQA & 10 & 128 & 0.0 & 0.0 \\
AOKVQA-MC & 15 & 128 & 0.0 & 0.0 \\
ActivityNet-QA & 1 & 128 & 1e-06 & 0.0 \\
MSRVTT-QA & 1 & 128 & 0.0 & 0.0 \\
MSVD-QA & 1 & 128 & 3e-07 & 0.0 \\
\bottomrule
\end{tabular}
\end{sc}
\end{small}
\end{center}
\vskip -0.1in
\end{table}

\paragraph{Tasks} 
We evaluated our approach on six image-based question-answering tasks from the CoIN benchmark \citep{chen2024coin}: VQAV2 \citep{goyal2017making}, TextVQA \citep{singh2019towards}, OCR-VQA \citep{mishra2019ocr}, ScienceQA \citep{lu2022learn}, VizWizVQA \citep{gurari2018vizwiz}, and GQA \citep{hudson2019gqa}. Following the sequences of tasks proposed by the authors, we also included an additional randomly shuffled order. The three task sequences are as follows:
\begin{itemize}
    \item ScienceQA, TextVQA, GQA, VizWizVQA, VQAv2, and OCR-VQA
    \item GQA, OCR-VQA, ScienceQA, TextVQA, VizWizVQA, and VQAV2
    \item TextVQA, OCR-VQA, VizWizVQA, GQA, ScienceQA, and VQAV2
\end{itemize}

\begin{figure}[t]
\vskip 0.2in
\begin{center}
\includegraphics[width=0.5\columnwidth]{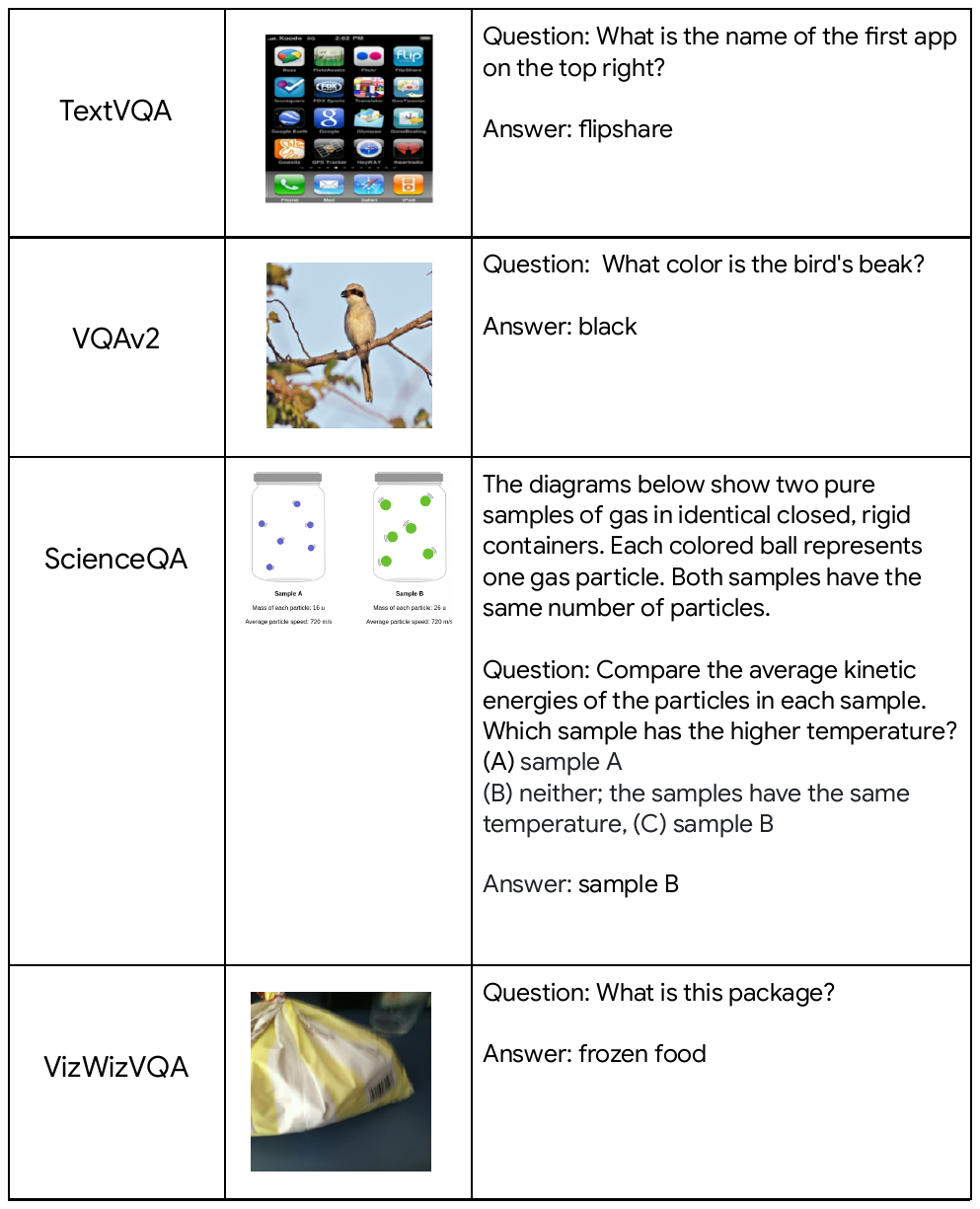}
\includegraphics[width=0.48\columnwidth]{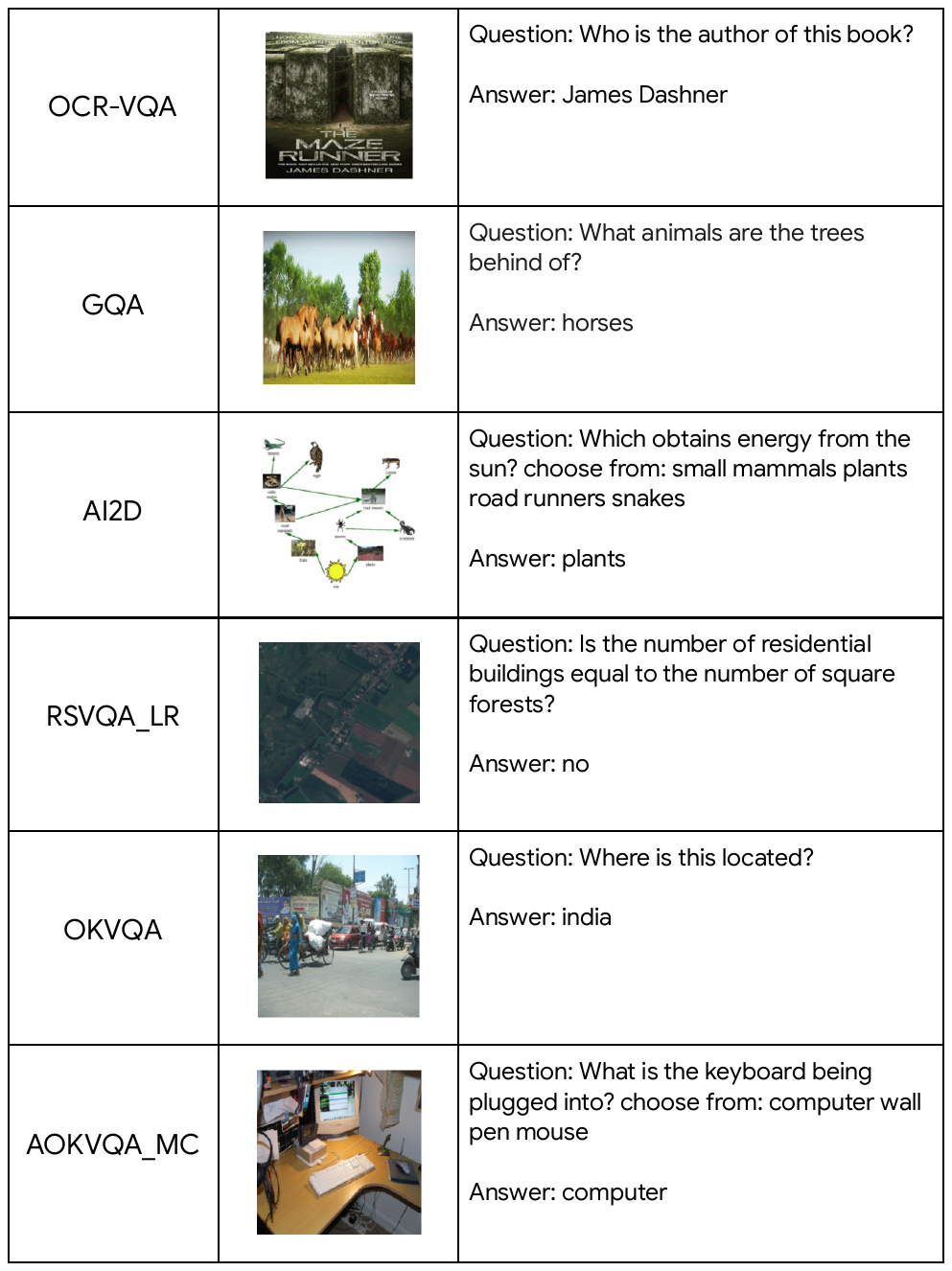}
\caption{Examples from each task in the extended CoIN benchmark.}
\label{coin_examples}
\end{center}
\end{figure}

We introduced an extended version of the CoIN benchmark by incorporating four additional tasks (AI2D \citep{kembhavi2016diagram}, AOKVQA-MC \cite{schwenk2022okvqa}, OKVQA \citep{marino2019ok}, and RSVQA-lr \citep{lobry2020rsvqa}), enhancing task diversity and creating a more challenging, longer sequence. The tasks were arranged in the following order: ScienceQA, TextVQA, AOKVQA-MC, OCR-VQA, RSVQA-lr, VizWizVQA, GQA, OKVQA, VQAv2, and AI2D. \cref{coin_examples} illustrates examples from each task. We found that ScienceQA is one of the most challenging tasks, often causing interference with others, likely due to its higher degree of dissimilarity. This task involves reasoning and answering multiple-choice questions on topics spanning natural science, language science, and social science.  

We also experiment with video question answering tasks, constructing a sequence of three tasks: MSVD-QA \citep{xu2017video}, MSRVTT-QA \citep{xu2017video}, and ActivityNet-QA \citep{yu2019activitynet}. 

\begin{table}[t]
\caption{Performance of PAM across three different run on a sequence of three tasks.}
\label{table:std}
\begin{center}
\begin{small}
\begin{sc}
\begin{tabular}{lcc}
\toprule
 & ACC & BWT \\
\midrule
PAM & 68.47$\pm$0.06 & -2.08$\pm$0.03 \\
\bottomrule
\end{tabular}
\end{sc}
\end{small}
\end{center}
\vskip -0.1in
\end{table}

\paragraph{Hyper-parameters} We used the best hyperparameter found by \citet{beyer2024paligemma} in fine-tuning each transfer task, expect for the learning rate, where we used a value of 5e-4. We report the hyper-parameters for all tasks in Table \ref{table:hyperparameters}. We used a LoRA of a rank 32 for all experiments and analyzed other values in \cref{fig:effect_lora_init_and_rank}. LoRA modules are added to attention and feedforward layers. We perform the alignment procedure every 100 steps, and use alignment percentage of 50\% in all experiments. We analyzed the effect of these hyper-parameters in Tables \ref{table:effect_align_percentage} and \ref{table:effect_of_alignment_schedule}. Our main results characterize performance variability as the standard deviation across different task orders. We used one seed for our experiments due to resource constraints and the high computational cost of learning a sequence of tasks with VLMs. This is motivated by findings in \citep{beyer2024paligemma} showing low standard deviation across runs for most transfer tasks. To further validate this for our setup, we performed multiple runs with three seeds on a sequence of three tasks and provided the results in \cref{table:std}. Consistent with previous findings, we observe that the standard deviation across three run is very small.    

\paragraph{Baselines} We used a LoRA rank of 32 for all baselines. For TIES. \citep{yadav2024ties}, we tuned the percentage of weights retained before alignment (i.e., during the pruning of redundant weights, as shown in Figure 1 in \citep{yadav2024ties}). We explored values from the grid [0.5, 0.7, 0.9] and found that 0.9 yielded the best results. TALL \citep{wanglocalizing} includes a hyperparameter that determines the amount of information the mask extracts from the multi-task vector. Lower values lead to more parameters being selected for a task. We searched over the grid [0.1, 0.2, 0.4] and found that 0.1 performed the best.

\begin{table}[t]
\caption{Effect of alignment schedule on PAM performance.}
\label{table:effect_of_alignment_schedule}
\begin{center}
\begin{small}
\begin{sc}
\resizebox{0.35\textwidth}{!}{%
\begin{tabular}{lccc}
\toprule
Schedule & ACC & BWT & $A_t$\\
\midrule
10 & 48.25 & -21.21 & 65.93\\
100 & 48.16 & -21.37 & 65.97 \\
300 & 47.82 & -21.78 & 65.97\\
\bottomrule
\end{tabular}
}
\end{sc}
\end{small}
\end{center}
\vskip -0.1in
\end{table}

\begin{figure}
\vskip 0.2in
\begin{center}
\includegraphics[width=0.45\textwidth]{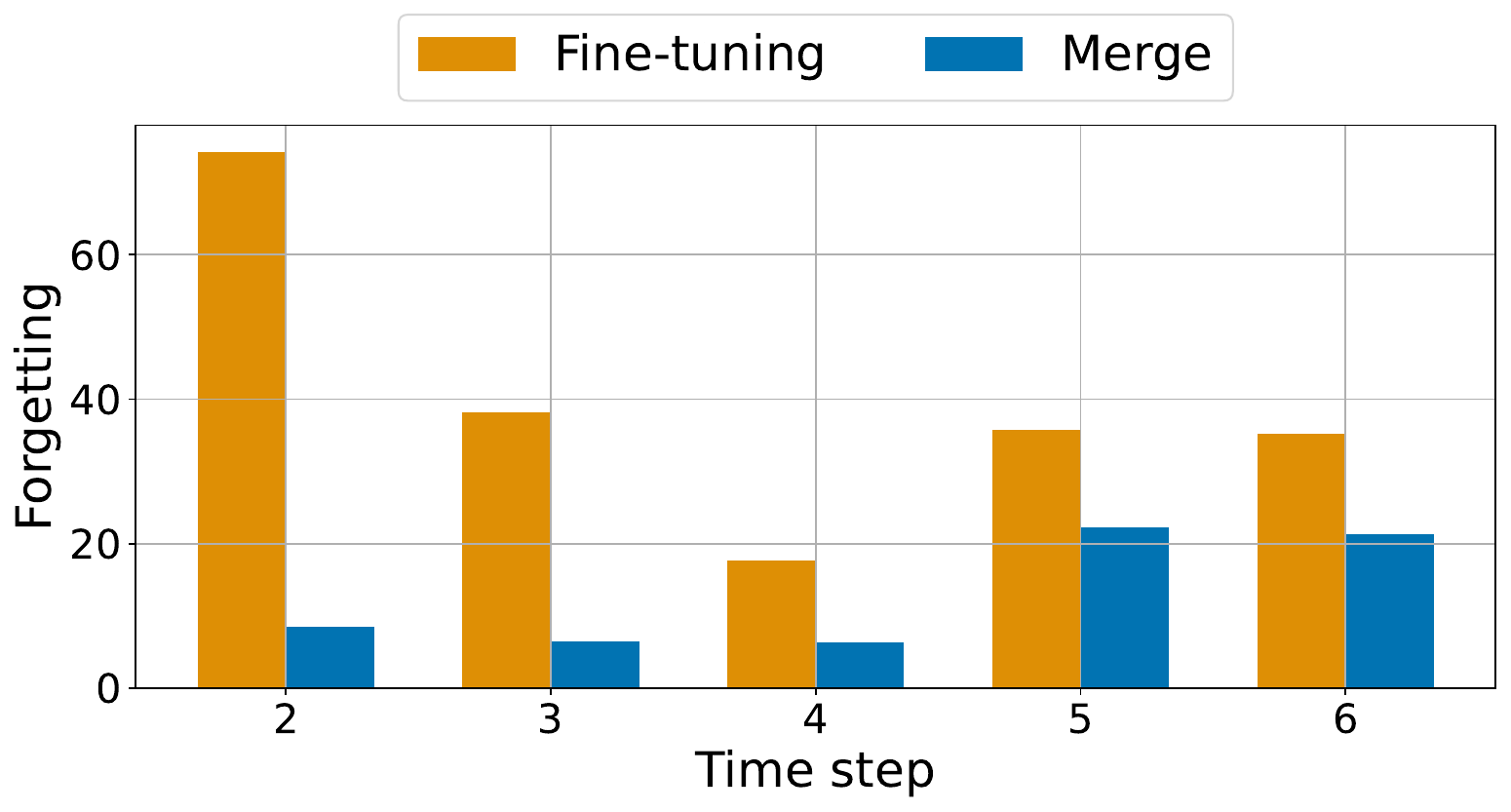} 
\includegraphics[width=0.45\textwidth]{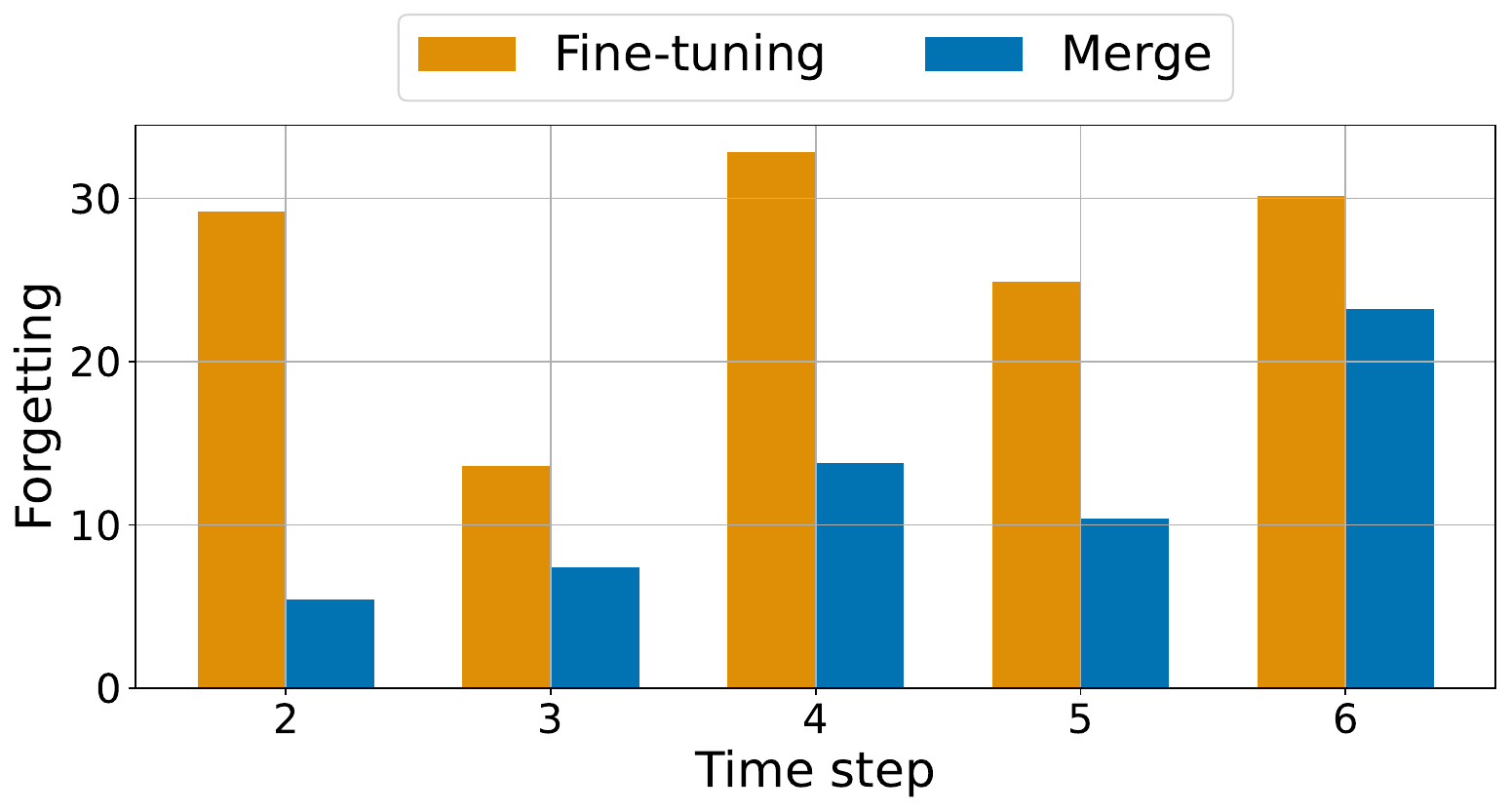}
\caption{Forgetting over time on the CoIN benchmark \citep{chen2024coin} for the second (left) and third (right) orders. At each time step, we report the average forgetting on seen tasks so far (lower is better).}
\label{fig:detailed_forgetting}
\end{center}
\vskip -0.2in
\end{figure}

\paragraph{Evaluation metrics} We do extensive evaluation by assessing four metrics to measure  model capabilities. 
\ \\
(1) \textit{Average accuracy (ACC)}: Let $A_{t,i}$ be the accuracy of task $i$ after learning task $t$. The average accuracy of all tasks at the end of the sequence of $T$ tasks is defined as:     
\begin{equation}
    ACC = \frac{1}{T}\sum_{i=1}^T A_{T,i}.
\end{equation}
(2) \textit{Backward transfer} (BWT) \citep{lopez2017gradient}: This metric measures the influence of learning task $t$ on the performance of each previous task $i < t$. Thus, it assess the model robustness against forgetting. BWT is defined as:
\begin{equation}
    BWT = \frac{1}{T-1}\sum_{i=1}^{T-1} A_{T,i} - A_{i,i}.
\end{equation}
(3) \textit{Forward transfer} (FWT) \citep{lopez2017gradient}: This metric measures the ability of a model to generalize to unseen tasks. It is estimated by the difference between the model performance at time step $t$ on unseen tasks ($i > t$) and the zero-shot performance. Specifically, let $\bar{a_{i}}$ be the zero-shot accuracy of task $i$, FWT is defined as:   

\begin{equation}
FWT = \frac{1}{T-1}\sum_{i=2}^{T-1} A_{i-1,i} - \bar{a_{i}}.   
\end{equation}

(4) \textit{Average task Accuracy} ($A_t$): this metric assess the plasticity of the model to learn a new task. It measures the performance of task $t$ immediately after learning it, and is defined as follows: 

\begin{equation}
A_t = \frac{1}{T}\sum_{i=1}^{T} A_{i,i} .  
\end{equation}

\label{appendix:extra_results}

\section{Additional Experiments}

\begin{figure}
\vskip 0.2in
\begin{center}
\includegraphics[width=0.4\columnwidth]{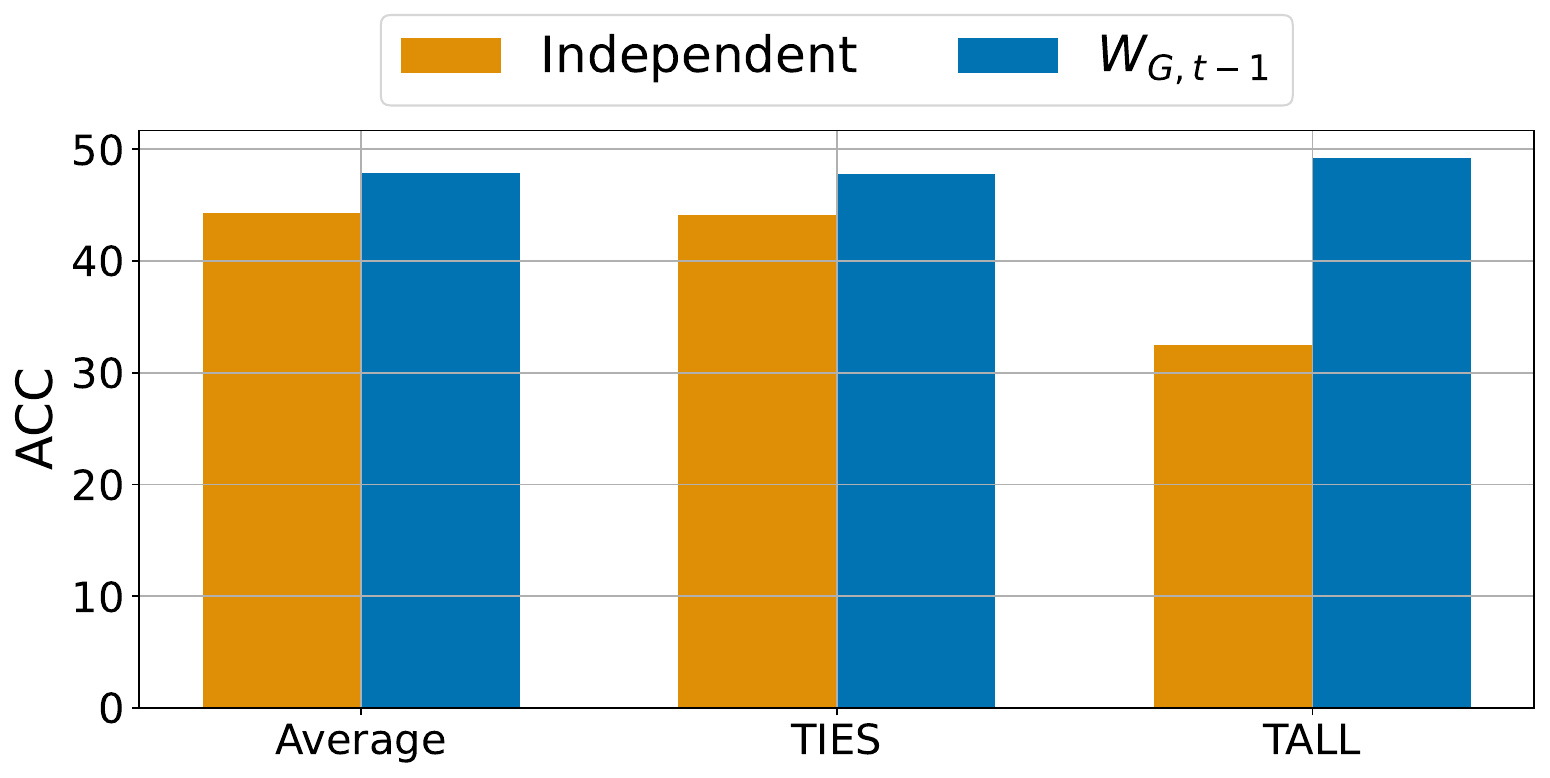}
\hspace{1cm}
\includegraphics[width=0.25\columnwidth]{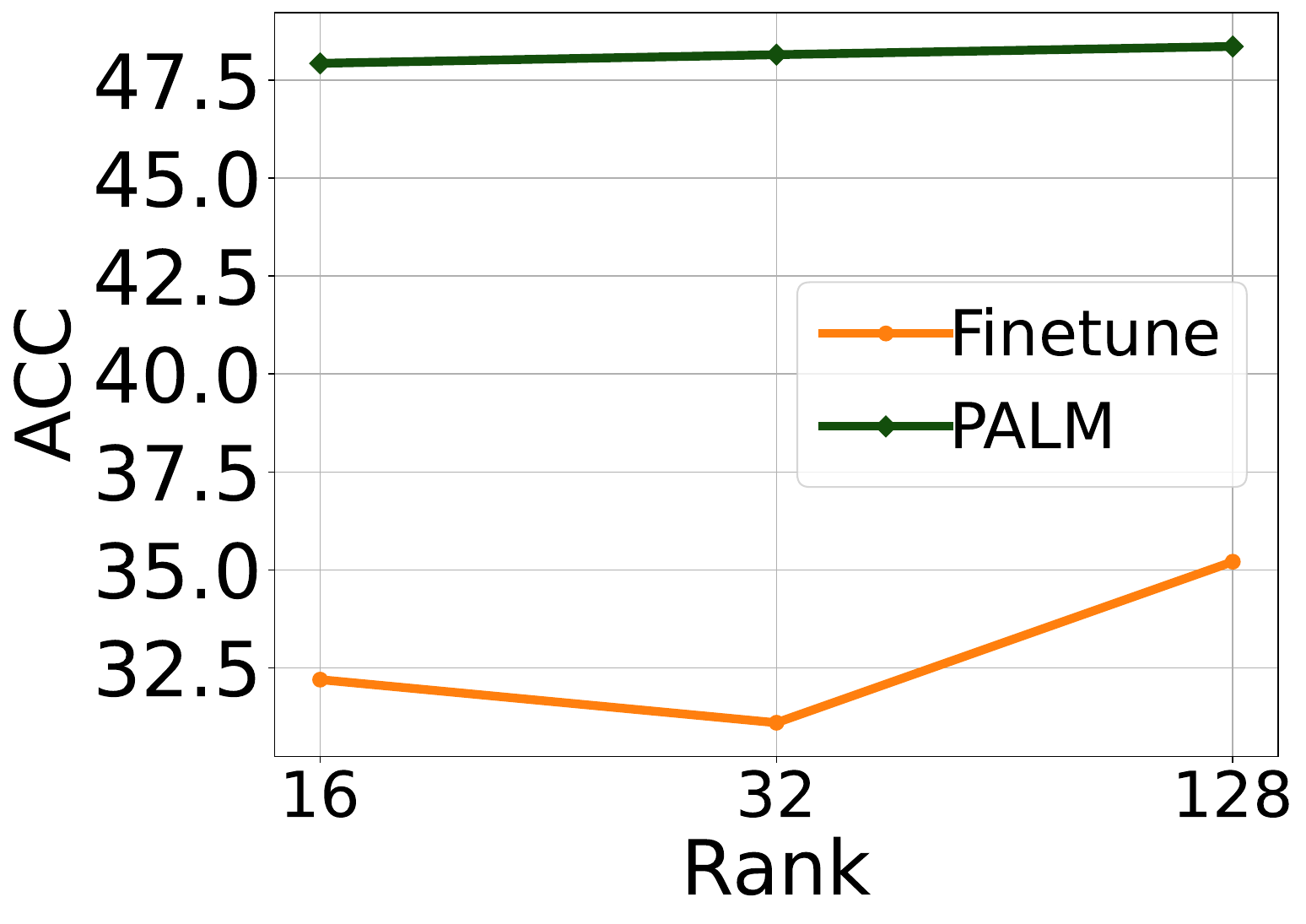}
\caption{Impact of weight initialization. Using global LoRA weights to initialize new task weights is more effective across different merging methods (left). Effect of LoRA rank on Performance. PAM can use low rank without scarifying performance, enabling more computational efficient training over time (right).}
\label{fig:effect_lora_init_and_rank}
\end{center}
\vskip -0.2in
\end{figure}

\paragraph{Detailed results on CoIN benchmark} Figure \ref{figure1} presents the forgetting over time for the standard order of CoIN. We illustrate the results for the other two considered orders in Figure \ref{fig:detailed_forgetting}. Consistent with previous observations, fine-tuning is more prone to forgetting and the performance of a task varies over time; some tasks have a greater impact on previous task performance than others, likely due to varying degrees of task similarity. In contrast, merging demonstrates greater robustness to these task variations.

\paragraph{Impact of LoRA initialization}
\label{sec:lora_initalization}
Merging methods typically operate on independently trained models. 
In the context of merging LoRAs, 
we compare performance of initializing new LoRA module with the global LoRA module weights, versus starting from an independent initialization. 
%
To isolate the effect of initialization, we analyze the performance of the average merging baseline without the periodic alignment process. 
As illustrated in \cref{fig:effect_lora_init_and_rank} left, initializing a new LoRA module from the global one leads to better performance compared to randomly initializing this module. We find that this observation is consistent across different merging algorithms including TIES and TALL. This finding highlights the importance of initialization strategies and provides insights for designing effective LoRA merging mechanisms.

\paragraph{Impact of LoRA Rank}
Previous studies have explored the impact of network width on continual learning performance, observing that wider networks generally exhibit reduced catastrophic forgetting \citep{mirzadeh2022wide}.
This phenomenon can be attributed to the increased capacity of wider networks, allowing them to accommodate new information without overwriting previously learned knowledge.
We investigated whether this trend holds true for LoRA, where the rank of the adapter can be considered to be analogous to network width.  While smaller ranks are more efficient, we hypothesized that they might exacerbate forgetting due to their limited capacity. 

Our results in \cref{fig:effect_lora_init_and_rank} right confirm that higher ranks (e.g., 128) lead to improved performance for fine-tuning. This suggests that increasing the rank provides additional capacity, similarly as in full fine-tuning, allowing the model to better retain previous knowledge. 
However, our proposed model, PAM, maintains strong performance even at low ranks, demonstrating its efficiency. This makes PAM a particularly attractive option for resource-constrained environments or applications where model size is a critical factor.

\end{document}